\documentclass[10pt]{llncs}
\usepackage{graphicx}
\usepackage{amsmath,amssymb} 
\usepackage{color}
\usepackage[width=122mm,left=12mm,paperwidth=146mm,height=193mm,top=12mm,paperheight=217mm]{geometry}
\begin{document}
\pagestyle{headings}
\mainmatter

\title{A Saccaded Visual Transformer for General Object Spotting}

\author{W.T.Pye, D.A.Sinclair}
\institute{Willempye@gmail.com, University of Warwick. \\david@imense.com, Imense Ltd. 
\\28th Oct 2022}

\maketitle

\begin{abstract}
  This paper presents the novel combination of a visual transformer style patch classifier with saccaded local attention.
  A novel optimisation paradigm for training object models is also presented, rather than the optimisation function
  minimising class membership probability error the network is trained to estimate the normalised distance
  to the centroid of labeled objects. This approach builds a degree of translational invariance directly
  into the model and allows fast saccaded search with gradient ascent to find object centroids. 
  The resulting saccaded visual transformer is demonstrated on human faces.
\end{abstract}

\section{Introduction}
\label{sec:intro}

The task of recognising or spotting general objects in images remains combinatorically difficult.
The space of visual appearance of objects is both large and scales geometrically with the number of pixels in
any image of an object to be recognised. Increasing the resolution of images of an object can make the
computational aspects of object recognition harder.

Smaller distinguished sub-parts can be used to reduce the relevant search space for refined model matching. 
This style of recogniser is exemplified by finding SIFT\cite{DBLP:journals/ijcv/Lowe04} or
MSER\cite{DBLP:conf/bmvc/MatasCUP02} etc. style features followed by geometric matching or bag of words indexing,
reviews of discrete feature detectors include \cite{gauglitz2011evaluation}.

Convolution nets have represented the state of the art in object classification \cite{726791} but deeper networks like Resnet50 \cite{https://doi.org/10.48550/arxiv.1512.03385} offer a greater capacity to learn larger scale objects.

The Visual Transformer \cite{DBLP:journals/corr/abs-2010-11929} offers a different approach using a higher dimensional model consisting of
embedded feature vectors derived from learned basis functions for sub-patches covering an extended image patch.
The size of the resulting network means large training sets are required (300M images to train 90M weights) and the networks are
slow to apply. Adding new models to a network requires significant training resources. The degree to which the ViT demonstrates
translational invariance is not clear.

This paper presents a novel optimisation paradigm for training transformer style object spotting/recognition networks
that allows fast saccaded search. Visual transformers are typically trained through presenting vast numbers of larger image patches
that contain single exemplars of visual categories to the training system. The attraction of this being a single recogniser
can be trained to distinguish thousands of visual categories at once. In the training process a set of basis functions
are learned that allow the creation of a set of linear coefficients to represent the content of sub-patches and their relative locations.
Visual inspection of the basis functions in the ViT paper \cite{DBLP:journals/corr/abs-2010-11929}
show them to bear at least a passing resemblance to
visual receptive fields in higher animals \cite{hubel1963shape}.

This paper uses a similar network structure but seeks to create translation tolerant single object recognisers through
the use of a novel optimisation criteria. Rather than training on error in probability of object class the system optimises
predicted scaled distance from the labeled object centroid. Additionaly sampling points may be drawn
from a subset of image locations which are of high visual interest.  
In this paper the visual interest operator chosen is the ST-transform \cite{DBLP:journals/corr/abs-2102-02000}, returning the
connected dark/light boundary chains as geometrically filterable candidate saccade points.
Geometric filtering of the dark/light boundary could follow the ideas in \cite{DBLP:conf/bmvc/RichardsFJ92} and \cite{DBLP:journals/cvgip/RichardsH85}.

The paper is layed out as follows: section \ref{sec:distance} details the \{bf Di\}stance to \{bf  F\}eature \{bf T\}raining function,
the PyTorch network definition of the model being given in appendix\ref{sec:pytorch}. 
Section \ref{sec:saccade} quantifies the speed benefits of only applying expensive classifiers at appropriate saccade points.
Conclusions an directions for future research are given in section \ref{sec:conclusions}.

\section{Distance to feature based training}
\label{sec:distance}

In this paper we introduce $distance to feature based training$ (\textbf{DiFT}) this involves sampling
training patches from an image and assigning a target score based on the distance between the centre
of the patch and the labeled centroid of a feature. 

We use the Large-scale CelebFaces Attributes (CelebA) Dataset\cite{liu2015faceattributes} as training data for our models.
The dataset used contains rectified images where faces are scaled to a standard size in a 178x218 colour jpeg, this sets
the scale for sample image patches and distance measures.
We constructed training patches by taking a pair of random coordinates, which satisfy a minimum distance
from the edge of the image ($\lfloor \frac{x}{2} \rfloor$ for a patch of size $x$).
Image patched can also be selected to be centered on candidate saccade points for an image.

\subsection{Target score method}
\label{subsec:score}

\begin{figure}[htbp]
    \centering
    \includegraphics[totalheight=2in]{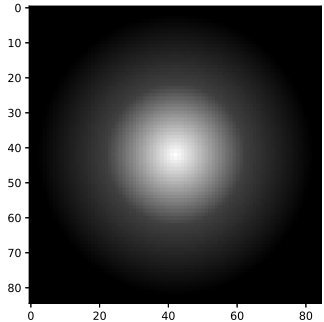}
    \caption{Plot of the score function, \ref{eq:score_func}, on an 85 by 85 pixel grid.
     White represents a score of one, and black a score of zero.}
    \label{fig:score_func}
\end{figure}

\begin{figure}[htbp]
    \centering
    \includegraphics[width=0.9\linewidth]{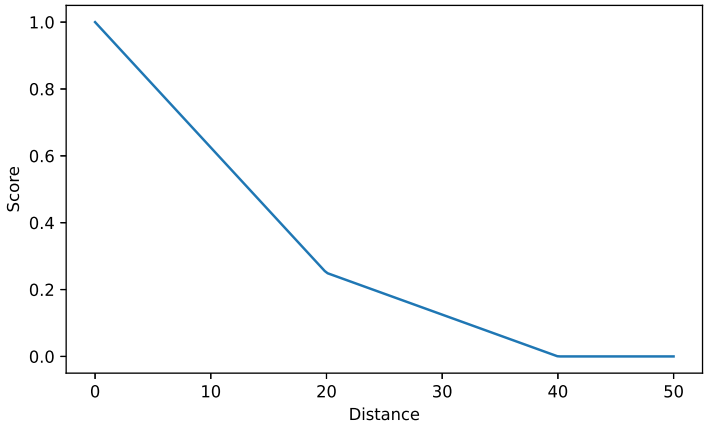}
    \caption{A graph of \ref{eq:score_func}, showing linearly decreasing score as distance increases until score is zero at forty pixels away from the feature}
    \label{fig:score_graph}
\end{figure}

Let $(x_{1},y_{1})$ be the point at the centre of the training patch, and let $(x_{2},y_{2})$ be the point representing the feature.
Then $D_{12}$ is defined as the Euclidean distance between the two points.
\begin{equation}
    D_{i2} = \sqrt{(x_{1} - x_{2})^{2} + (y_{1} - y_{2})^{2}}
    \label{eq:dist}
\end{equation}
We then assign the training patch a target score between zero and one as a function of distance such that $score = f(D)$; for the function $f$ we used \ref{eq:score_func} as shown below.

\begin{equation}
    f(x) = 
    \left\{
        \begin{array}{lr}
            0 & \quad x > 40 \\
            \frac{1}{2} -\ \frac{x}{80} & \quad 40 \geq x > 20 \\
            1 - \frac{3x}{80} & \quad 20 \geq x \geq 0 
        \end{array}
    \right.
    \label{eq:score_func}
\end{equation}
This piece-wise function decreases from $1$ to $0.25$ as $x$ increases from $0$ to $20$, then decreases linearly to $0$ as $x$ increases to $40$, with $0$ gradient beyond $x = 40$. 
$f$ is shown in \ref{fig:score_func} and in \ref{fig:score_graph}

In the case of an eye spotter image samples taken more than $40$ pixels form a labeled eye are guaranteed to contain stuff that is not an eye.

\begin{figure}[htbp]
    \centering
    \includegraphics[width=0.9\linewidth]{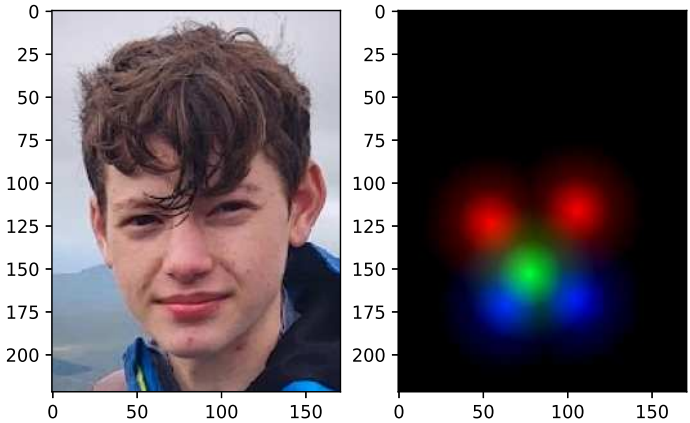}
    \caption{The score function applied to several features on a face. The red fields are centred on the centre of the eye, the green field on the tip of the nose and the blue fields on the mouth corners}
    \label{fig:target_face}
\end{figure}

\begin{figure}[htbp]
    \centering
    \includegraphics[width=1\linewidth]{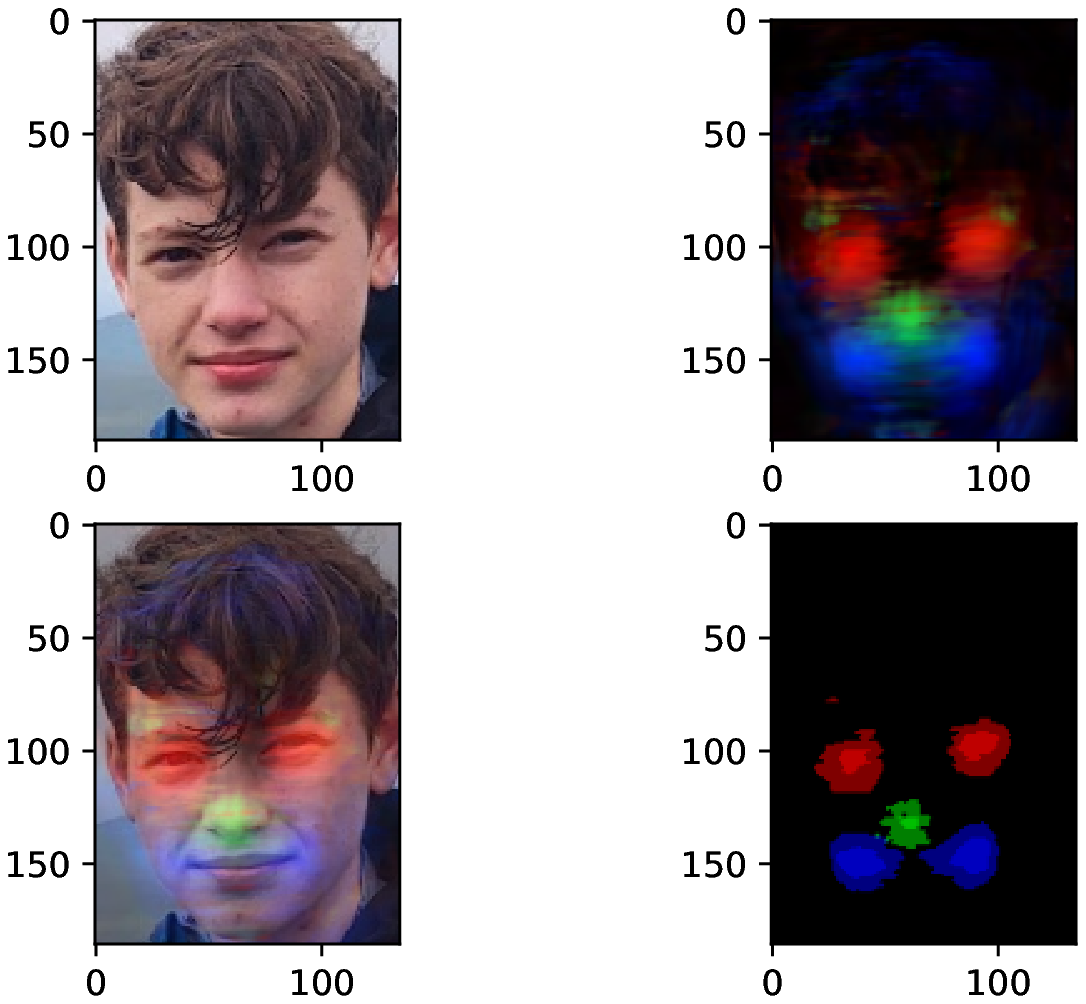}
    \caption{Distance to feature based training applied to a CNN on a 35 by 35 patch.
    Red shows predicted distance from eyes, green for nose and blue for mouth corners.
    Top left is the input, top right is the output, bottom left is the output overlaid on the input, bottom right is the output with values below 0.5 set to 0, 0.5 -- 0.8 set to 0.5 and above 0.8 set to 0.8.}
    \label{fig:face_features}
\end{figure}

This distance to feature method can easily be extended to multiple features by using a separate channel for each feature, as can be seen in \ref{fig:target_face} where one channel is used for eyes, another for nose tips and a third for mouth corners.
This is how we trained a model on the landmarks from the CelebA dataset.
The lower left image of \ref{fig:face_features} was produced by applying a convolutional neural network (CNN)\cite{6795724} with a 35 by 35 pixel patch input trained as described above and then applied to the original image pixel-wise.

\subsection{Benefits of distance to feature based training}
\label{subsec:DiFTbenefits}

The first benefit of distance to feature based training is that it can massively reduce the effort in creating training data.
By assigning each pixel within a labeled image a score, each pixel effectively becomes a piece of training data, allowing one labeled image to become potentially thousands of training patches.
This can reduce the amount of labeled training data needed and lower the impact of over-fitting models.
Another benefit is that \textbf{ DiFT} facilitated the use of gradient descent based methods to find feature centroids.

\subsection{Heatmaps and basis functions}
\label{sec:basis}

\begin{figure}[htbp]
    \centering
    \includegraphics[width=\linewidth]{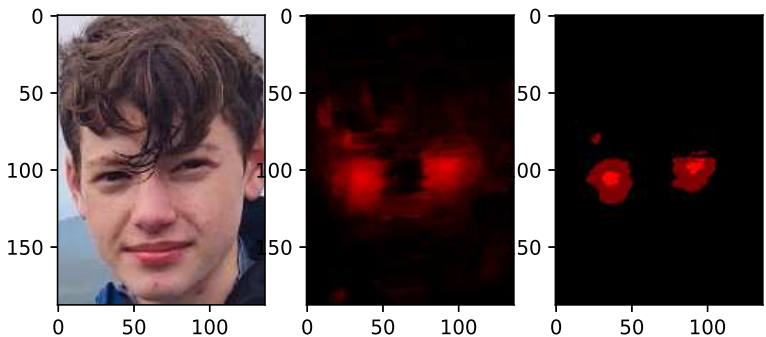}
    \caption{Image transformed by applying a CNN over a 31 by 31 pixel patch pixel-wise over the image.}
    \label{fig:eye_transform}
\end{figure}

Figure \ref{fig:eye_transform} shows the estimated distance to feature returned by the $visual transformer$ style CNN defined in \ref{}.
As can be seen the most eye-like locations are centered on the eyes.

Innate to the $visual transformer$ \cite{DBLP:journals/corr/abs-2010-11929} style of network is the linear
embedding space which is effectively a set of learned convolution kernels. These learned convolution kernels arguably play a
roll similar to receptive fields in the human visual system. While the ViT convolution kernels are not generally orthogonal
they can be used as basis vectors in a weighted reconstruction of a sub-patch. From casual inspection
the basis vectors perform poorly at reconstructing edge detail. This is perhaps a pointer to a
limitation in the ability of the ViT to represent detailed shape.

\begin{figure}[htbp]
    \centering
    \includegraphics[width=\linewidth]{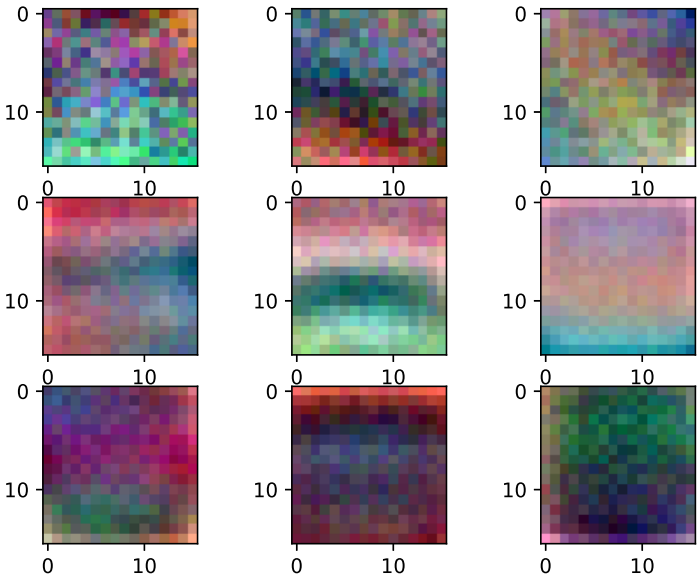}
    \caption{The 16 by 16 convolution kernels of the CNN used in \ref{fig:eye_transform}}
    \label{fig:eye_kernels}
\end{figure}

A limited number of convolution kernels (basis vectors for feature embedding)
was used in this paper largely because of limited training resources
(one Mac mini M1 running PyTorch).

\subsection{The role of Saccade in fast visual processing}
\label{sec:saccade}

\begin{figure}[htbp]
    \centering
    \includegraphics[width=0.4\linewidth]{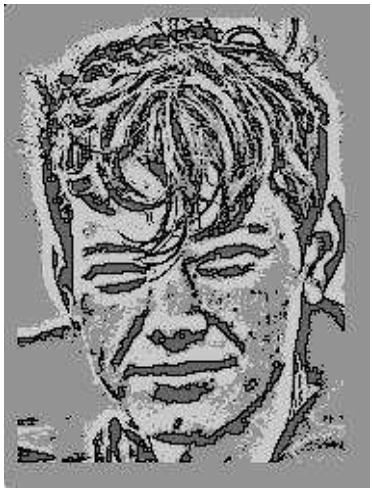}
    \caption{ST-transform applied to a face image, black pixels are dark/light region boundaires.
      There are significantly fewer boundary pixels than image pixels}
    \label{fig:saccade}
\end{figure}

Saccade is a quintessential part of the human visual system \cite{nature_saccade}.
The human eye interogates a scene via a series of rapid directed shifts ($saccades$) to point at 
feature points which provide an instantaneously stabilised retinal image. Even when the eye has
alighted on a particular spot micro-saccades continue. Rather than uniformly scanning an
image for objects to be recognised it is sufficient to focus on interest points. 
In this paper we choose to use the darl/light boundary chains returned by the ST
transform\cite{DBLP:journals/corr/abs-2102-02000} to provide a set of interest points.
The length scale of the model together with the property that it returns an estimate of the distance to 
a model centroid means that only a fraction of the generalised feature points need to be sampled
for object detection.
Figure\ref{fig:saccade} shows the result of the ST-transform applied to a face image. The face image has 46,000
pixels giving dark/light boundary chains of 5,200 pixels which could be crudely sampled every 5 pixels to give
1,040 candidate search locations for objects of interest. Other geometry based boundary filters could be
applied to give candidate saccade points.

\section{Conclusions}
\label{sec:conclusions}

Using the magnitude of distance from a sampled image patch to a labeled object centroid as the thing to be
optimised in a Visual Transformer style network provides a classifier that is robust to translation with
in the scale of the model. Using a generalised feature to provide candidate saccade points for image search
vastly reduces processing time for visual search.

\section{Acknowledgements}
\label{sec:acknowledgements}

Mr L. Pye for permitting the use of his face in this publication.
If you wish to reproduce his image please contact \texttt{lucasbpye@gmail.com}.
Imense Ltd for an interesting summer internship.

To cite: \cite{DBLP:journals/corr/VaswaniSPUJGKP17}, \cite{Comaniciu02meanshift:}


\bibliographystyle{splncs}
\bibliography{egbib}

\begin{thebibliography}{10}

\bibitem{DBLP:journals/ijcv/Lowe04}
Lowe, D.G.:
\newblock Distinctive image features from scale-invariant keypoints.
\newblock Int. J. Comput. Vis. \textbf{60}(2) (2004)  91--110

\bibitem{DBLP:conf/bmvc/MatasCUP02}
Matas, J., Chum, O., Urban, M., Pajdla, T.:
\newblock Robust wide baseline stereo from maximally stable extremal regions.
\newblock In Rosin, P.L., Marshall, A.D., eds.: Proceedings of the British
  Machine Vision Conference 2002, {BMVC} 2002, Cardiff, UK, 2-5 September 2002,
  British Machine Vision Association (2002)  1--10

\bibitem{gauglitz2011evaluation}
Gauglitz, S., H{\"o}llerer, T., Turk, M.:
\newblock Evaluation of interest point detectors and feature descriptors for
  visual tracking.
\newblock International journal of computer vision \textbf{94}(3) (2011)
  335--360

\bibitem{726791}
Lecun, Y., Bottou, L., Bengio, Y., Haffner, P.:
\newblock Gradient-based learning applied to document recognition.
\newblock Proceedings of the IEEE \textbf{86}(11) (1998)  2278--2324

\bibitem{https://doi.org/10.48550/arxiv.1512.03385}
He, K., Zhang, X., Ren, S., Sun, J.:
\newblock Deep residual learning for image recognition (2015)

\bibitem{DBLP:journals/corr/abs-2010-11929}
Dosovitskiy, A., Beyer, L., Kolesnikov, A., Weissenborn, D., Zhai, X.,
  Unterthiner, T., Dehghani, M., Minderer, M., Heigold, G., Gelly, S.,
  Uszkoreit, J., Houlsby, N.:
\newblock An image is worth 16x16 words: Transformers for image recognition at
  scale.
\newblock CoRR \textbf{abs/2010.11929} (2020)

\bibitem{hubel1963shape}
Hubel, D.H., Wiesel, T.N.:
\newblock Shape and arrangement of columns in cat's striate cortex.
\newblock The Journal of physiology \textbf{165}(3) (1963)  559

\bibitem{DBLP:journals/corr/abs-2102-02000}
Sinclair, D., Town, C.:
\newblock A generalised feature for low level vision.
\newblock CoRR \textbf{abs/2102.02000} (2021)

\bibitem{DBLP:conf/bmvc/RichardsFJ92}
Richards, W., Feldman, J., Jepson, A.D.:
\newblock From features to perceptual categories.
\newblock In Hogg, D.C., Boyle, R., eds.: Proceedings of the British Machine
  Vision Conference, {BMVC} 1992, Leeds, UK, September, 1992, {BMVA} Press
  (1992)  1--10

\bibitem{DBLP:journals/cvgip/RichardsH85}
Richards, W., Hoffman, D.D.:
\newblock Codon constraints on closed 2d shapes.
\newblock Comput. Vis. Graph. Image Process. \textbf{31}(3) (1985)  265--281

\bibitem{liu2015faceattributes}
Liu, Z., Luo, P., Wang, X., Tang, X.:
\newblock Deep learning face attributes in the wild.
\newblock In: Proceedings of International Conference on Computer Vision
  (ICCV). (December 2015)

\bibitem{6795724}
LeCun, Y., Boser, B., Denker, J.S., Henderson, D., Howard, R.E., Hubbard, W.,
  Jackel, L.D.:
\newblock Backpropagation applied to handwritten zip code recognition.
\newblock Neural Computation \textbf{1}(4) (1989)  541--551

\bibitem{nature_saccade}
Gaarder, K.:
\newblock Transmission of edge information in the human visual system.
\newblock Nature \textbf{212} (1966)

\bibitem{DBLP:journals/corr/VaswaniSPUJGKP17}
Vaswani, A., Shazeer, N., Parmar, N., Uszkoreit, J., Jones, L., Gomez, A.N.,
  Kaiser, L., Polosukhin, I.:
\newblock Attention is all you need.
\newblock CoRR \textbf{abs/1706.03762} (2017)

\bibitem{Comaniciu02meanshift:}
Comaniciu, D., Meer, P.:
\newblock Mean shift: A robust approach toward feature space analysis.
\newblock In: In PAMI. (2002)  603--619

\bibitem{PyTorch_url}
:
\newblock Pytorch: An imperative style, high-performance deep learning library

\end{thebibliography}

\appendix

\section{PyTorch Network Definition}
\label{sec:pytorch}

PyTorch \cite{PyTorch_url} was used as the NN training environment. A series of networks
were evaluated with the one used to generate the results in this paper having the
following definition within PyTorch.

\begin{verbatim}

class conv_face_features(nn.Module):
    def __init__(self, drop=0.):
        super().__init__()
        self.conv1 = nn.Conv2d(3,9,16)
        self.conv2 = nn.Conv2d(9,18,11)
        self.linear1 = nn.Linear(100,50)
        self.linear2 = nn.Linear(50*18, 256)
        self.linear3 = nn.Linear(256,64)
        self.linear4 = nn.Linear(64,16)
        self.linear5 = nn.Linear(16,3)
        self.act = nn.Mish()
        self.drop = nn.Dropout(p=drop)
        
        
    def forward(self,x):
        # x is torch.tensor(dtype=torch.float32) batch of patches [batchsize,3,height,width]
        x = x / 255.
        x = x.permute(0,3,1,2)
        x = self.conv1(x)
        
        x = self.drop(x)
        x = self.act(x)
        x = self.conv2(x)
        x = x.flatten(-2,-1)
        
        x = self.drop(x)
        x = self.act(x)
        x = self.linear1(x)
        x = x.flatten(-2,-1)
        
        x = self.drop(x)
        x = self.act(x)
        x = self.linear2(x)
        
        x = self.drop(x)
        x = self.act(x)
        x = self.linear3(x)
        
        x = self.drop(x)
        x = self.act(x)
        x = self.linear4(x)
        
        x = self.drop(x)
        x = self.act(x)
        x = self.linear5(x)
        return x

def training2(net, batches, batchsize, lr=0.05, momentum=0.9):
    net.train()
    criterion = nn.MSELoss() #MSE works, Cross-Entropy didnt work.
    optimizer = torch.optim.SGD(net.parameters(), lr=lr, momentum=momentum)
    running_loss = 0.
    for i in range(batches):
        d = specsfreeimage()
        inputs = torch.empty((batchsize,31,31,3), dtype=torch.float32)
        scores = torch.empty(batchsize, dtype=torch.float32)
        for j in range(batchsize): #All patches in a batch are from same image to reduce time spent loading images.
            r = rand_coords(border=30)
            mindist = min_dist(d,r)
            scores[j] = score(mindist[0],1)
            inputs[j] = torch.from_numpy(patch(d,r)).to(torch.float32)
        optimizer.zero_grad()
        outputs = net(inputs)
        loss = criterion(outputs, scores.unsqueeze(1))
        loss.backward()
        optimizer.step()
        running_loss += loss.item()
        print(i, ': ', str(running_loss/(i+1))[:7])
    
\end{verbatim}

\end{document}